\documentclass{article}

 \PassOptionsToPackage{numbers, compress}{natbib}

\usepackage[final]{neurips_2024}

\usepackage[english]{babel}
\usepackage[controls]{animate}
\usepackage{tikz}
\usetikzlibrary{arrows}
\usepackage{algorithm}
\usepackage{algpseudocode}
\usepackage{amsmath}
\usepackage{amsfonts}
\usepackage{graphicx}
\usepackage[colorlinks=true, allcolors=blue]{hyperref}
\usepackage{booktabs}
\usepackage{cleveref}

\newcommand{\first}[1]{\textbf{\textcolor{red}{#1}}}
\newcommand{\second}[1]{\textbf{\textcolor{violet}{#1}}}
\newcommand{\third}[1]{\textbf{\textcolor{black}{#1}}}

\usepackage{amsmath,amsfonts,bm}

\def\eqref#1{equation~\ref{#1}}
\def\Eqref#1{Equation~\ref{#1}}

\def\1{\bm{1}}

\DeclareMathAlphabet{\mathsfit}{\encodingdefault}{\sfdefault}{m}{sl}
\SetMathAlphabet{\mathsfit}{bold}{\encodingdefault}{\sfdefault}{bx}{n}

\newcommand{\bfA}{{\bf A}}

\newcommand{\bfD}{{\bf D}}

\newcommand{\bfI}{{\bf I}}

\newcommand{\bfQ}{{\bf Q}}
\newcommand{\bfR}{{\bf R}}

\newcommand{\bfU}{{\bf U}}

\newcommand{\bfx}{{\bf x}}

\newcommand{\bfu}{{\bf u}}
\newcommand{\bfq}{{\bf q}}

\newcommand{\grad}{{\boldsymbol \nabla}}

\newcommand{\bftheta}{{\boldsymbol \theta}}

\newcommand{\bfeta}{{\boldsymbol \eta}}

\newcommand{\bfkappa}{{\boldsymbol \kappa}}

\renewcommand{\grad}{{\boldsymbol \nabla\, }}

\usepackage[utf8]{inputenc} 
\usepackage[T1]{fontenc}    
\usepackage{hyperref}       
\usepackage{url}            
\usepackage{booktabs}       
\usepackage{amsfonts}       
\usepackage{nicefrac}       
\usepackage{microtype}      
\usepackage{xcolor}         

\title{Advection Augmented Convolutional Neural Networks}

\author{%
  Niloufar Zakariaei \\
  University of British Columbia\\
      Vancouver, Canada \\
      \texttt{nilouzk@student.ubc.ca} \\
  \And 
  Siddharth Rout \\ 
    University of British Columbia\\
        Vancouver, Canada \\
            \texttt{siddharth.rout@ubc.ca} \\
  \And 
Eldad Haber \\ 
    University of British Columbia\\
    Vancouver, Canada \\
    \texttt{ehaber@eoas.ubc.ca} \\
  \And 
Moshe Eliasof \\ 
University of Cambridge \\
Cambridge, United Kingdom\\
\texttt{me532@cam.ac.uk}
  \\
}

\begin{document}

\maketitle

\begin{abstract}
Many problems in physical sciences are characterized by the prediction of space-time sequences. Such problems range from weather prediction to the analysis of disease propagation and video prediction. Modern techniques for the solution of these problems typically combine Convolution Neural Networks (CNN) architecture with a time prediction mechanism. However, oftentimes, such approaches underperform in the long-range propagation of information and lack explainability. In this work, we introduce a physically inspired architecture for the solution of such problems. Namely, we propose to augment CNNs with advection by designing a novel semi-Lagrangian push operator. We show that the proposed operator allows for the non-local transformation of information compared with standard convolutional kernels. We then complement it with Reaction and Diffusion neural components to form a network that mimics the Reaction-Advection-Diffusion equation, in high dimensions. We demonstrate the effectiveness of our network on a number of spatio-temporal datasets that show their merit.
\end{abstract}

\section{Introduction and Motivation}

Convolution Neural Networks (CNNs) have long been established as one of the most fundamental and powerful family of algorithms for image and video processing tasks, in applications that range from image classification \cite{krizhevsky2012imagenet, he2016deep}, denoising \cite{burger2012image}  and reconstruction \cite{kingma2013auto}, to generative models \cite{goodfellow2014generative}. More examples of the impact of CNNs on various fields and applications can be found in \cite{o2015introduction, gu2018recent, li2021survey} and references within.

At the core of CNNs, stands the convolution operation -- a simple
linear operation that is local and spatially rotation and translation equivariant.
The locality of the convolution, coupled with
nonlinear activation functions and deep architectures
have been the force driving CNN architectures to the forefront of machine learning and artificial intelligence research \cite{simonyan2014very, he2016deep}.
One way to understand the success of CNNs and attempt to generate an explainable framework to them is to
view CNNs from a Partial Differential Equation (PDE)
point of view \cite{RuthottoHaber2018, chen2018neural}. In this framework, the convolution is viewed as a mix of discretized differential operators of varying order. The layers of the network are then associated with time. Hence, the deep network can be thought of as a discretization of a nonlinear time-dependent PDE. Such observations have motivated 
parabolic network design that smooth and denoise images \cite{ruthotto2019deep} as well as to  
networks that are based on hyperbolic equation \cite{lensink2022fully} and semi-implicit architectures \cite{haber2019imexnet}.

However, it is known from the literature \cite{li2020multipole}, and is also demonstrated in our experiments, that CNN architectures tend
to under-perform in tasks that require rapid
transportation (also known as \emph{advection}) of information from one side of 
an image to the other. In particular, in this paper, we 
focus on the prediction of the spatio-temporal behavior of image features, where significant transportation is present in the data.
Examples of such data include the prediction of weather, 
traffic flow, and crowd movement.

{\bf Related work:} In recent years, significant research was devoted for addressing spatio-temporal
problems. Most of the works known to us are built on a combination of CNN to capture spatial dependencies and Recurrent Neural Networks (RNN) to capture temporal 
dependencies. A sample of papers that address this problem and the related problem of video prediction can be found in \cite{bohnstingl2022online, tan2022simvp, seo2023implicit, li2023moganet, hsieh2018learning, mathieu2015deep, finn2016unsupervised}   and reference within. See also \cite{wikle2019comparison} and \cite{oprea2020review} for a recent comparison between different methods. Such methods typically behave as black boxes, in the sense that while they offer strong downstream performance, they often times lack a profound understanding of the learned underlying dynamics from the data. 
Another type of work that is designed for the scientific datasets is \cite{takamoto2023pdebench}, which uses Fourier-based methods to build the operators. See also \cite{blechschmidt2021three} for a review on the topic.

\begin{figure}
\centering
\begin{tabular}[t]{ccc}
    \begin{tikzpicture}
  \draw[step=.5cm] (-1.5,-1.5) grid (1.5,1.5);
  \filldraw[draw=red,fill=lightgray] (-1.0,-1.0) rectangle (-1.5,-1.5);
\end{tikzpicture} &
\begin{tikzpicture}
    \draw[step=.5cm] (3.499,-1.5) grid (6.5,1.5);
  \filldraw[draw=red,fill=lightgray] (6,1) rectangle (6.5,1.5);
\end{tikzpicture}
&
\includegraphics[width=4.28cm]{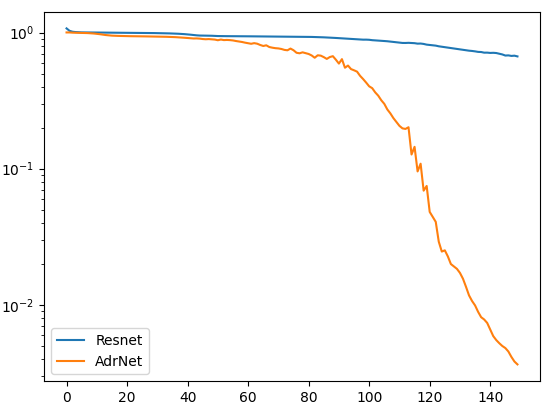} \\
(A) Source image  & (B) Target image & (C) Convergence

\end{tabular}
\caption{A simple task of moving information from one side of the image to the other. The source image in A is moved to the target image in B. The convergence of a simple ResNet and an ADRnet proposed in this work is in (C). 
\label{fig1}} 
\end{figure}
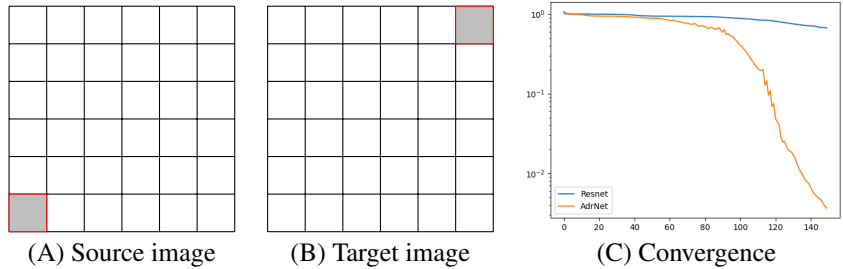

{\bf Motivation:} Notably, while a CNN is a versatile tool that allows to learn spatial dependencies, it can have significant challenges in learning simple operations that require transportation.
As an example, let us consider the problem of predicting the motion in the simple case that the input data is
an image, where all pixels take the value of $0$ except for a pixel on the bottom left (marked in gray), and the
output is an image where the value is transported to a pixel on the top right. This example is illustrated in Figure~\ref{fig1}.
Clearly, no local operation, for example, a convolution of say, $3 \times 3$  or even $7 \times 7$ can be used to move the information from the bottom left of the image to the top right. Therefore, the architecture to achieve this task requires either many convolutions layers, or, downsampling the image via pooling, where the operations are local, performing convolutions on the downsampled image, and then upsampling the image via unpooling, followed by additional convolutions to "clean" coarsening and interpolation artifacts, as is typical in UNets \cite{UNET2015, Unet3D}. To demonstrate, we attempt to fit the data with a simple convolution residual network and with a residual network that has an advection block, as discussed in this paper. The convergence history for the two methods is plotted in Figure~\ref{fig1}. We see that while a residual network is incapable of fitting the data, adding an advection block allows it to fit the data to machine precision.

This set of problems, as well as the relatively poor performance it offers on data that contains advection as in simple task in Figure \ref{fig1} sets the motivation for our work.
Our aim is to extend the set of tools that is available in CNNs
beyond simple and local convolutions. For time-dependent PDEs, it is well known that it is possible to model most phenomena by a set of advection-diffusion-reaction equations (see, e.g., \cite{EvansPDE, eliasof2023adr} and references within). Motivated by the connection between the discretization of PDEs and deep network \cite{RuthottoHaber2018, chen2018neural}, and our observations on the shortcomings of existing operations in CNNs, we propose
reformulating CNNs into three different components. Namely, (i) a pointwise term, also known as a \emph{reaction} term, where channels interact locally. (ii) A \emph{diffusion} term, where features are exchanged between neighboring pixels in a smooth manner. And, (iii) an \emph{advection} term, where features are passed from pixels to other pixels, potentially not only among neighboring pixels, while preserving \emph{feature mass or color loss}\footnote{That is the sum of the features is constant.}. As we discuss in Section~\ref{sec:method}, the combination of diffusion and reaction is equivalent to a standard CNN. However, there is
no CNN mechanism that is equivalent to the advection term. Introducing this new term allows the network flexibility in cases where information is carried directly.

\paragraph{Contributions:} \label{contri} The contributions of this paper are three-fold. First, we form the 
spatio-temporal dynamics in high dimensions as an advection-diffusion-reaction process, which is novel and has not been studied in CNNs prior to our work. Second, we propose the use of the semi-Lagrangian approach for its solution, introducing a new type of a learnable linear layer, that is sparse yet non-local. This is in contrast to standard convolutional layers, which act locally. In contrast to advection, other mechanisms for non-local interactions,
require dense interactions, which are computationally expensive \cite{vaswani2017attention}. Specifically, our use of semi-Lagrangian methods offers a bridge
between particle-based methods and convolutions \cite{LentineGretarssonFedkiw2011}. Thus, we
present a new operation in the context of CNNs, that we call the 
\emph{push operator} to implement the advection term. This operator allows us to transport features anywhere on the image in a single step -- an operation that cannot be modeled with small local convolution kernels. It is thus a simple yet efficient replacement to the standard techniques that are used to move information on an image.
Third, we propose a methodology to learn these layers based on the splitting operator approach, and show that they can successfully model advective processes that appear in different datasets. 

\paragraph{Limitations:}
The advection diffusion reaction model is optimal when applied to the prediction of images where the information for the prediction is somehow present in the given images. Such scenarios are often present in scientific applications. For example, for the prediction of the propagation of fluids or gasses, all we need to know is the state of the fluid now (and in some cases, in a few earlier time frames). A more complex scenario is the prediction of video. In this case, the next frame may have new features that were not present in previous frames. To this end, the prediction of video requires some generative power. While we show that our network can be used for video prediction and even obtain close to the state-of-the-art results, we observe that it performs best for scientific datasets. 

\section{Model Formulation}

\paragraph{Notations and assumptions.}
We consider a spatio-temporal vector function of the form $\bfq(t, \bfx) = [\bfq_1(t, \bfx), \ldots, \bfq_m(t, \bfx)] \in {\cal Q}$, where $\cal Q$  is the space vector function with $m$ channels. The function $\bfq$ is defined over the domain
 $\bfx \in \Omega \subseteq \mathcal{R}^{d}$, and time interval $[0,t_j]$.
Our goal is to predict the function at time $t_k$ for some $t_k>t_j$, given the inputs up to time $j$. For the problem we consider here, the time is sampled on a uniform grid with equal spacing. Below, we define the advection-diffusion-reaction system that renders the blueprint of the method proposed in this paper to achieve our goal.

\paragraph{Reaction-Advection-Diffusion System.} Given  the input function $\bfq$, we first embed it in a higher dimensional space. We denote the embedding function by $\bfI: \in {\cal I}$, defined as
\begin{eqnarray}
    \label{embedding}
    \bfI(t,\bfx) = M_{\rm{In}}(\bfq(t,\bfx), \bftheta_{\rm{In}})
\end{eqnarray}
where $M_{\rm{In}}:\mathbb{R}^{m} \rightarrow \mathbb{R}^{c}$ is a multi-layer preceptron (MLP) that embeds
the function $\bfq$ from $m$ to $c>m$ channels with trainable parameters $\bftheta_{\rm{In}}$.

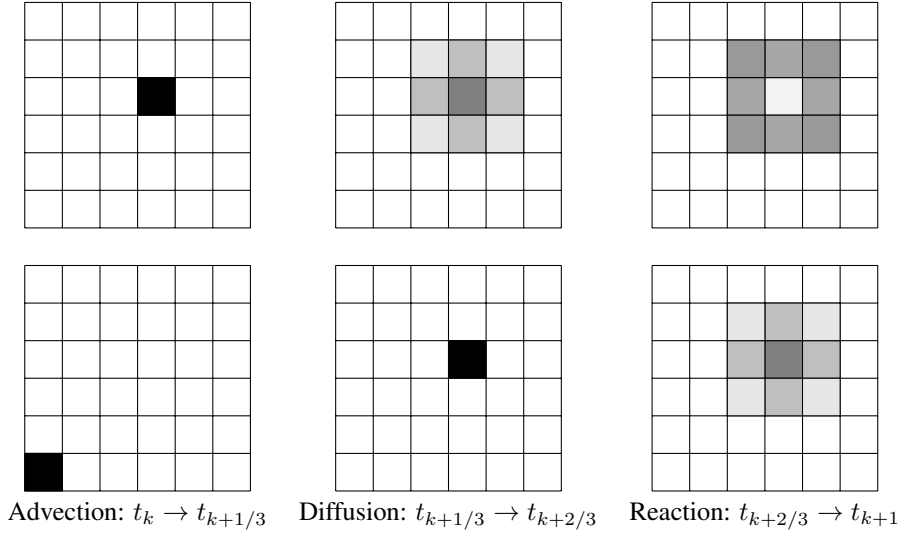
\begin{figure}
\centering
\begin{tabular}{ccc}
    \begin{tikzpicture}
    
  \draw[step=0.5cm] (-1.5,-1.5) grid (1.5,1.5);
  \filldraw[fill=black] (-1.5,-1.5) rectangle (-1,-1);

\draw[step=0.5cm] (-1.5,1.999) grid (1.5,5.0);
  \filldraw[fill=black] (0,3.5) rectangle (0.5,4.0);
\end{tikzpicture} &  
    \begin{tikzpicture}
  \draw[step=0.5cm] (-1.5,-1.5) grid (1.5,1.5);
\filldraw[fill=black] (0, 0) rectangle (0.5,0.5);

\draw[step=0.5cm] (-1.5,1.999) grid (1.5,5.0);
  \filldraw[fill=gray] (0, 3.5) rectangle (0.5,4.0);

  \filldraw[fill=lightgray] (-0.5, 3.5) rectangle (0,4.0);
  \filldraw[fill=lightgray] (0.5, 3.5) rectangle (1,4.0);
  \filldraw[fill=lightgray] (0, 3.0) rectangle (0.5,3.5);
  \filldraw[fill=lightgray] (0, 4.0) rectangle (0.5,4.5);
  \filldraw[fill= gray!20] (-0.5, 3.0) rectangle (0,3.5);
  \filldraw[fill=gray!20] (-0.5, 4) rectangle (0,4.5);
  \filldraw[fill=gray!20] (0.5, 3.0) rectangle (1,3.5);
  \filldraw[fill=gray!20] (0.5, 4) rectangle (1,4.5);

\end{tikzpicture}
& 
    \begin{tikzpicture}
  \draw[step=0.5cm] (-1.5,-1.5) grid (1.5,1.5);

  \filldraw[fill=gray] (0, 0) rectangle (0.5,0.5);
   \filldraw[fill=lightgray] (-0.5, 0) rectangle (0,0.5);
  \filldraw[fill=lightgray] (0.5, 0) rectangle (1,0.5);
  \filldraw[fill=lightgray] (0, -0.5) rectangle (0.5,0.0);
  \filldraw[fill=lightgray] (0, 0.5) rectangle (0.5,1);
  \filldraw[fill= gray!20] (-0.5, -0.5) rectangle (0,0.0);
  \filldraw[fill=gray!20] (-0.5, 0.5) rectangle (0,1);
  \filldraw[fill=gray!20] (0.5, 0.5) rectangle (1,1);
  \filldraw[fill=gray!20] (0.5, -0.5) rectangle (1,0);

\draw[step=0.5cm] (-1.5,1.999) grid (1.5,5.0);

 \filldraw[fill=gray!10] (0, 3.5) rectangle (0.5,4.0);
   \filldraw[fill=gray!70] (-0.5, 3.5) rectangle (0,4.0);
  \filldraw[fill=gray!70] (0.5, 3.5) rectangle (1,4.0);
  \filldraw[fill=gray!70] (0, 3.0) rectangle (0.5,3.5);
  \filldraw[fill=gray!70] (0, 4.0) rectangle (0.5,4.5);
  \filldraw[fill= gray!80] (-0.5, 3.0) rectangle (0,3.5);
  \filldraw[fill=gray!80] (-0.5, 4.0) rectangle (0,4.5);
  \filldraw[fill=gray!80] (0.5, 4.0) rectangle (1,4.5);
  \filldraw[fill=gray!80] (0.5, 3.0) rectangle (1,3.5);
 
  \end{tikzpicture} \\
  Advection: $t_k \rightarrow t_{k+1/3}$ &  Diffusion:   $t_{k+1/3} \rightarrow t_{k+2/3}$ & Reaction: $t_{k+2/3} \rightarrow t_{k+1}$
\end{tabular}
\caption{An illustration of the advection-diffusion reaction process. In the first step (advection), a pixel on the lower left of the image is transported into the middle of the mesh. In the second step (diffusion), the information is diffused to its neighbours and finally, in the last step (reaction) each pixel interact locally to change its value. \label{fig:process_illustration}}

\end{figure}

To represent the evolution of $\bfq$ we evolve $\bfI$
in the hidden dimension, $c$, and then project it back into the space ${\cal Q}$.
One useful way to represent the evolution of a spatio-temporal process is by combining three
different processes, as follows:
\begin{itemize}
    \item \emph{Reaction:} A pointwise process where channels interact pointwise (sometimes referred to as $1\times 1$ convolutions)
    \item \emph{Diffusion:} A process where features are being communicated and diffused locally.
    \item  \emph{Advection:} A process where information transports along mediums.
\end{itemize}
These three processes are also illustrated in Figure \ref{fig:process_illustration} and their composition defines the advection-diffusion-reaction differential equation
on the embedded vector $\bfI$.

The equation can be written as
\begin{eqnarray}
    \label{eq:adr}
    {\frac {\partial \bfI(t,\bfx)}{\partial t}} &=& \kappa \Delta \bfI(t,\bfx) + \nabla \cdot \left( \bfU \bfI(t,\bfx)\right) + R(\bfI(t,\bfx), \bftheta), \\
    \bfI(t=0,\bfx) &=& M(\bfq(t=0,\bfx)).
\end{eqnarray}
Here $\Delta$ is the Laplacian and $\nabla$ is the divergence operator, as classically defined in PDEs \cite{EvansPDE}. The equation is equipped with an initial condition and some boundary conditions. Here, for simplicity of implementation, we choose the Neumann boundary conditions, but other boundary conditions can also be chosen. 
The diffusivity coefficient $\kappa$, velocity field $\bfU$, and the parameters that control the reaction term $R$ are trainable, and are discussed in Section \ref{sec:method}.

The equation is integrated on some interval $[0,T]$ and finally one obtains $\bfq(T,\bfx)$ by applying
a second MLP, that projects the hidden features in $\bfI(t=T, \bfx)$ to the desired output dimension, which in our case is the same as the input dimension, i.e., $m$:
\begin{eqnarray}
    \label{embeddingb}
    \bfq(T,\bfx) = M_{\rm{Out}}(\bfI(T,\bfx), \bftheta_{\rm{Out}}),
\end{eqnarray}
where $\bftheta_{\rm{out}}$ are trainable parameters for the projection MLP.

\textbf{Remark (\Eqref{eq:adr} Reformulation).} The discretization of \Eqref{eq:adr} can be challenging due to conservation properties of the term $\nabla \cdot \left( \bfU \bfI(t,\bfx)\right)$. An alternative equation, which may be easier to discretize in our context, can be obtained by noting that
\begin{eqnarray}
    \label{eq:mass-color}
     {\frac {\partial \bfI(t,\bfx)}{\partial t}} + \nabla \cdot (\bfU \bfI) =  {\frac {\partial \bfI(t,\bfx)}{\partial t}} + \bfU \cdot \nabla \bfI +
    \bfI \nabla \cdot \bfU. 
\end{eqnarray}
The operator on the left-hand side in \Eqref{eq:mass-color} is the continuity equation \cite{EvansPDE}, where the {\em mass} of $\bfI$ is conserved. The first two terms on the right hand side, namely, $\bfI_t + \bfU \cdot \nabla \bfI$ are sometimes refer to as the color equation \cite{EvansPDE} as they conserve the {\em intensity} of $\bfI$. For divergent free velocity fields, that is, when $\grad \cdot \bfU = 0$, these are equivalent, however, for non-divergent fields, the term $\bfI  \nabla \cdot \bfU$ is a pointwise operator on $\bfI$, that is, it is a reaction term. When training a model, one can use either \Eqref{eq:adr} in its continuity form or replace the term with \Eqref{eq:mass-color} and learn the term  $\bfI  \nabla \cdot \bfU$ as a part of the reaction term, $R$. We discuss this in discretization of our model in Section \ref{sec:reaction}.

\section{From a Partial Differential Equation to a Neural Network}
\label{sec:method}

To formulate a neural network from the differential equation in \Eqref{eq:adr} needs to be discretized in time and space.
In this work, we assume data that resides on a regular, structured mesh grid, such as 2D images, and the spatial operators to discretize \Eqref{eq:adr} are described below.
To discretize \Eqref{eq:adr} in \emph{time}, we turn to Operator Splitting methods \cite{ascherBook} that are common
for the discretization of equations with similar structures, and were shown to be effective in deep learning frameworks \cite{eliasof2023adr}.
As we see next, such discretization leads to a neural
a network that has three types of layers that are composed of each other, resulting in an effectively deeper neural network.

\subsection{Operator Splitting}

The idea behind operator splitting is to split the integration of the ODE into parts \cite{leveque}. Specifically, consider a linear differential equation of the form
\begin{equation}  
\label{eq:linear_adr}
 {\frac {\partial \bfI(t,\bfx)}{\partial t}} = \bfA \bfI(t,\bfx) + \bfD \bfI(t,\bfx) + \bfR \bfI(t,\bfx),
 \end{equation}
where $\bfA, \bfD$ and $\bfR$ are matrices.
The solution to this system at time $t$ is well known \cite{EvansPDE} and reads
\begin{equation}
\label{eq:solution_linear}
 \bfI(t, \bfx) = \exp \left(t \bfA + t \bfD + t \bfR \right)) \bfI(0, \bfx),  
 \end{equation}
where $\exp$ denotes the matrix exponentiation operation.
It is also possible to approximate the exact solution presented in \Eqref{eq:solution_linear} as follows
\begin{eqnarray}
\label{eq:os} \exp \left(t \bfA + t \bfD + t \bfR \right)) \bfI(0, \bfx) \approx   \exp (t \bfA)\left(( \exp (t \bfD) (\exp (t \bfR) \bfI(0, \bfx)) \right)
\end{eqnarray}
The approximation is of order $t$, and it stems from the fact that the eigenvalues of the matrices $\bfA, \bfD$ and $\bfR$ do not commute (see \cite{ascherBook} for a thorough discussion). 
 \Eqref{eq:os} can also be interpreted in the following way. The solution, for a short time integration time $t$, can be approximated by first solving the system
$ {\frac {\partial \bfI(t,\bfx)}{\partial t}} = \bfR \bfI(t, \bfx), \bfI_0 = \bfI(0, \bfx)$ obtaining a solution
$ \bfI_R(t, \bfx)$, followed by the solution of the system
$ {\frac {\partial \bfI(t,\bfx)}{\partial t}} = \bfD \bfI_{R}(t, \bfx), \bfI_0 = \bfI_R$ obtaining the solution $\bfI_{RD}(t, \bfx)$ and finally solving
the system $\frac {\partial \bfI(t,\bfx)}{\partial t} = \bfA \bfI_{RD}(t, \bfx), \bfI_0 = \bfI_{RD}$.
The advantage of this approach is that it allows the use
of different techniques for the solution of different problems. 

Let ${\cal R}$ be the solution operator that advances $\bfI(t_j,\bfx)$ to $\bfI_R(t_{j+1},\bfx)$. Similarly, 
let ${\cal D}$ be the solution operator that advances $\bfI_R(t_{j+1},\bfx)$ to $\bfI_{RD}(t_{j+1},\bfx)$
and lastly, let ${\cal A}$ be the solution of the advection problem that advances $\bfI_{RD}(t_{j+1},\bfx)$
to $\bfI(t_{j+1},\bfx)$. Then, a layer in the system can be written as the composite of three-layer
\begin{eqnarray}
    \label{adros}
  {\cal L}\, \bfI(t_j,\bfx) =   {\cal A}  \circ {\cal D} \circ {\cal R}\, \bfI(t_j,\bfx).
\end{eqnarray}
That is, the resulting discretization in time yields a neural network architecture of a layer that is composed of three distinct parts. We now discuss each part separately.

\subsection{Advection}

\begin{figure}
   \centering
   \usetikzlibrary {arrows.meta}
    \begin{tikzpicture}[scale=0.8]
  \draw[step=0.5cm] (-2.5,-2.5) grid (2.5,2.5);
  \filldraw[draw=red,fill=lightgray] (-2.5,-2.5) rectangle (-2,-2);
  
  \draw [-{Stealth[length=5mm]}] (-2.25,-2.25) -- (2.1,0.9);

  \draw [-{Stealth[length=2mm]}] (2.1,0.9) -- (1.75,1.25);
  \draw [-{Stealth[length=2mm]}] (2.1,0.9) -- (2.25,1.25);
  \draw [-{Stealth[length=2mm]}] (2.1,0.9) -- (2.25,0.75);
  \draw [-{Stealth[length=2mm]}] (2.1,0.9) -- (1.75,0.75);

  \node[black]  at (-2.25,-2.25) {$\bullet$};
  
  \node[red]  at (1.75,1.25) {$\bullet$};
  \node[red]  at (2.25,1.25) {$\bullet$};
  \node[red]  at (2.25,0.75) {$\bullet$};
  \node[red]  at (1.75,0.75) {$\bullet$};

  \draw[step=0.5cm] (3.499,-2.5) grid (8.5,2.5);
  \filldraw[draw=red,fill=lightgray] (7.5,1) rectangle (8.0,1.5);
\draw [-{Stealth[length=5mm]}] 
(4.1,-2.1) -- (7.75,1.25);
  \draw [-{Stealth[length=2mm]}] (4.25,-2.25) -- (4.1,-2.1);
  \draw [-{Stealth[length=2mm]}]  (4.25,-1.75) -- (4.1,-2.1);
  \draw [-{Stealth[length=2mm]}]  (3.75,-1.75) -- (4.1,-2.1);
  \draw [-{Stealth[length=2mm]}] (3.75,-2.25) -- (4.1,-2.1) ;

  \node[black]  at (7.75,1.25) {$\bullet$};
  \node[red]  at (4.25,-2.25) {$\bullet$};
  \node[red]  at (4.25,-1.75) {$\bullet$};
  \node[red]  at (3.75,-1.75) {$\bullet$};
  \node[red]  at (3.75,-2.25) {$\bullet$};

\end{tikzpicture}
  \caption{Discretization of the push operator. (a) Left: Semi-Lagrangian mass preserving transport, discretizing the continuity. (b) Right: Semi-Lagrangian color preserving transport. \label{fig2}}
\end{figure}
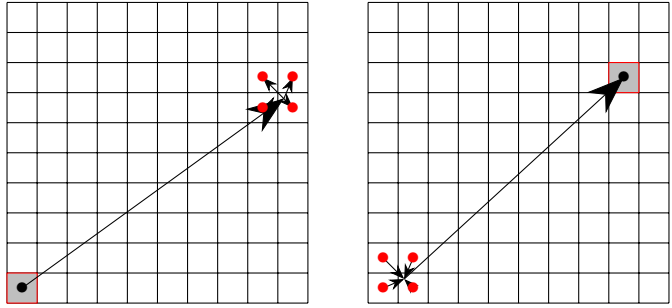

The innovative part of our network is advection.
The advection approximately solves the equation
\begin{eqnarray}
{\frac {\partial \bfI}{\partial t}} = \nabla \cdot \left( \bfU(\bfI,\bfx,t) \bfI \right),  
\end{eqnarray}
for a general velocity field $\bfU$. 
For the solution of this equation, 
we now introduce a linear operation that we use to enhance the performance of our network.
Our goal is to allow for information to pass over large distances. To this end, consider a displacement field
$\bfU = (\bfU_1, \bfU_2)$ and consider the push operation, $\bfA(\bfU) \bfI$ as the operation that takes
every pixel in $\bfI$ and displaces it from point $\bfx$
to $\bfx_u = \bfx+\bfU$. Since the point $\bfx_u$ does not necessarily reside on a grid point, the information
from $\bfx_u$ is spread over four grid points neighbors, in
weights that are proportional to the distance from these points. A sketch of this process in plotted in Figure~\ref{fig2} (a).
The operator discussed above conserves that {\em mass} of the features. A different implementation, as discussed in Remark 1, is to discretize the color equation. This is done by looking backward and using the interpolated value as shown in Figure~\ref{fig2}(b). It is possible to show \cite{FohringHaberRuthutto2013} that these linear operators are transposed of each other. Here, for each implementation, we chose to use the color equation. We show in ablation studies that the results when using either formulation are equivalent. 

The process allows for a different displacement vector $\bfu$ for every grid point. 
The displacement field $\bfU$ in has $2c$ channels and
can vary in space and time. To model the displacement field, we propose to use the data at times, 
\begin{eqnarray}
\label{Qk}
\bfQ_k = [\bfq(t_{k-j}, \bfx), \bfq(t_{k-j+1}, \bfx), \ldots, \bfq(t_{k}, \bfx)],
\end{eqnarray}
 where $j$ is the length
of history used to learn the displacements.

Using $\bfQ_k$, the displacement field is computed by
a simple residual convolution network, which we formally
write as
\begin{eqnarray}
    \label{disp}
    \bfU_k = RN(\bfQ_k, \bfeta),
\end{eqnarray}
where $RN$ is the residual network parameterized by $\bfeta$.

\subsection{Reaction}
\label{sec:reaction}
The reaction term is a nonlinear $1\times 1$ convolution. 
This yields a residual network of the form
\begin{eqnarray}
    \label{reaction}
    \bfI_{j+1} = \bfI_j + h M(\bfI_j, \bftheta_j) = {\cal R}_j(\bftheta_j)\, \bfI_j,
\end{eqnarray}
where $M$ is a standard, double-layer MLP with parameters 
$\bftheta_j$ and $h$ is a step size that is a hyper-parameter.
We may choose to have more than a single reaction step per iteration.

\subsection{Diffusion}

For the diffusion step, we need to discretize the Laplacian on the image. We use the standard 5-point Laplacian \cite{gvl} that can also be expressed as 2D group convolution \cite{nagyHansenBook}. 
Let $\Delta_h$
be the discrete Laplacian. The diffusion equation reads
$$ \bfI_{j+1} - \bfI_j = h \kappa\Delta_h \bfI_k.$$
If we choose $k=j$ we obtain an explicit scheme
\begin{eqnarray}
\label{fe}
    \bfI_{j+1} = \bfI_j + h \kappa\Delta_h \bfI_j.
\end{eqnarray}
Note that the diffusion layer can be thought of as a group convolution where each channel is convolved with the
same convolution and then scaled with a different $\kappa$. The forward Euler method for the diffusion
requires $h\kappa$ to be small if we want to retain stability.
By choosing $k=j+1$ we obtain the
backward Euler method, which is unconditionally stable
\begin{eqnarray}
\label{be}
    \bfI_{j+1} = (\bfI - h \kappa\Delta_h)^{-1} \bfI_j = {\cal D}(\kappa) \bfI_j.
\end{eqnarray}
To invert the matrix we use the cosine transform \cite{kelley1} which yields an $n \log n$ complexity
for this step.

\paragraph{Combining Diffusion and Reaction to a Single Layer.}
In the above network the diffusion is handled by an implicit method (that is a matrix inversion) and the reaction is handled by an explicit method. For datasets where the diffusion is significant, this may be important; however, in many datasets where the diffusion is very small, it is possible to use an explicit method for the diffusion.
Furthermore, since both the diffusion and reaction are computed by convolutions, it is possible to combine them
into a $3\times3$ convolution (see \cite{RuthottoHaber2018}
and \cite{haber2019imexnet} for a complete discussion).
This yields a structure that is very similar to a classical 
Convolutional Residual Network that replaces the diffusion and reaction steps. For the datasets used in this paper, we noted that this modest architecture was sufficient
to obtain results that were close to state-of-the-art.

\subsection{Implementing the ADR Network}

Implementing the diffusion and reaction terms, either jointly or combined, we use a standard Convolutional Residual Network.
The advection term is implemented by using the {\tt sampleGrid}
command in pytoch, which uses an efficient library to interpolate the images.

While the network can be used as described above, we found that better results can be obtained by denoising the output of the network. To this end, we have used a standard UNet and applied it to the output. As we show in our numerical experiments, this allows us to further improve downstream performance.
The complete network is summarized in Algorithm~\ref{alg:alg1}.
\begin{algorithm}
\caption{The ADR network}
\label{alg:alg1}
\begin{algorithmic}
\State Set $\bfI_0 \leftarrow M(\bfq_k, \bftheta_o)$, $\bfQ_k$
as in \eqref{Qk}.
\For{$j=0, 1, ...m-1$} 
    \State Diffusion-Reaction $\bfI_{DR} \leftarrow {\cal D}_{\bfkappa_j}   {\cal R}_{\bftheta_j} \bfI_j$
    \State Compute displacement $\bfU_j = RN(\bfI_{DR}, \bfeta_j)$  as in \eqref{disp}
    \State Push the image $\bfI_{j+1} = {\cal A}(\bfU_j) \bfI_{DR}$
\EndFor
\State Set $\bfq_{k+\ell} = M(\bfI_{m}, \bftheta_T)$
\State (Optional) Denoise $\bfq_{k+\ell} = {\rm Unet}(\bfq_{k+\ell})$
\end{algorithmic}
\end{algorithm}

\section{Experiments}
\label{expt}

Our goal is to develop architectures that perform well for scientific-related datasets that require advection. In our experiments, we use two such datasets, CloudCast \cite{zhang2017deep},
and the Shallow Water Equation in PDEbench \cite{takamoto2023pdebench}. However, our ADRNet can also be used for the solution of video prediction. While such problems behave differently than scientific datasets, we show that our ADRNet can perform reasonably well for those applications as well. Below, we elaborate on the utilized datasets. 
We run our codes using a single NVIDIA RTX-A6000 GPU with 48GB of memory.

\subsection{Datasets}

We now describe the datasets considered in our experiments, which are categorized below. 

\textbf{Scientific Datasets:} We consider the following datasets which arise from scientific problems and communities:
\begin{itemize}

\item 
\textbf{SWE}
The shallow-water equations are derived from the compressible Navier-Stokes equations.
The data is comprised of 900 sets of 101 images, each of which is a time step. 
    \item \textbf{CloudCast.}
The CloudCast dataset comprises 70,080 satellite images captured every 15 minutes and has a resolution of $3712 \times 3712$ pixels, covering the entire disk of Earth.
\end{itemize}

\textbf{Video Prediction Datasets:} These datasets are mainly from the Computer Vision community, where the goal is to predict future frames in videos. The datasets are as follows:
\begin{itemize}
    \item 
\textbf{Moving MNIST}
The Moving MNIST dataset is a synthetic video dataset designed to test sequence prediction models. It features 20-frame sequences where two MNIST digits move with random trajectories.

\item 
\textbf{KITTI}
The KITTI is a widely recognized dataset extensively used in mobile robotics and autonomous driving, and it also serves as a benchmark for computer vision algorithms. 
\end{itemize}

The statistics of the datasets are summarized in Table~\ref{tab:dataset},  and in Appendix \ref{app:exp}, we provide results on additional datasets, namely TaxiBJ \cite{zhang2017deep} and KTH \cite{schuldt2004recognizing}.
\begin{table}[h]
\centering
\caption{Datasets statistics. Training and testing splits, image sequences, and resolutions}
\vspace{10pt}
\begin{tabular}{lccccc}
\toprule
\textbf{Dataset} & \( N_{\text{train}} \) & \( N_{\text{test}} \) & \( (C, H, W) \) & \( History \) & \( Prediction \) \\
\midrule
PDEBench-SWE           & 900                  & 100                & (1, 128, 128)     & 10      & 1       \\
CloudCast        & 5241                  & 1741                 & (1, 128, 128)   & 4       & 4, 8, 12, 16        \\
Moving MNIST           & 10000                  & 10000                & (1, 64, 64)     & 10      & 10       \\
KITTI & 2042                 & 1983                 & (3, 154, 512)   & 2      & 1, 3        \\

\bottomrule
\end{tabular}
\label{tab:dataset}
\end{table}

\subsection{Evaluation}
\label{results}
\paragraph{Ranking of Methods.} Throughout all experiments where other methods are considered, we rank the top 3 methods using the color scheme of \first{First}, \second{Second}, and \third{Third}. 
\paragraph{Performance on Scientific Datasets.}
We start our comparisons with the SWE and CloudCast datasets. These datasets fit the description of our ADRNet as future images depend on the history alone (that is, the history should be sufficient to recover the future). Indeed, Table \ref{tab:comparisonSWE} and Table \ref{tab:cloudcast} show that our ADRNet performs much better than other networks for these goals.  
\begin{table}[h]
\footnotesize
\centering

\begin{minipage}{0.45\textwidth}
    \centering
        \caption{Results on PDEBench SWE Dataset.}
\begin{tabular}{lcccccc}
\toprule
Method & NRMSE $\downarrow$ \\
\midrule
UNET \cite{takamoto2022pdebench} & 8.3e-2 \\
PINN \cite{takamoto2022pdebench} & 1.7e-2 \\
MPP-AVIT-TI \cite{mccabe2023multiple} & 6.6e-3 \\
ORCA-SWIN-B \cite{shen2023orca} & 6.0e-3 \\
FNO \cite{takamoto2022pdebench} & 4.4e-3 \\
MPP-AVIT-B \cite{mccabe2023multiple} & \third{2.4e-3} \\
MPP-AVIT-L \cite{mccabe2023multiple} & \second{2.2e-3} \\
\midrule
{ADRNet} & \first{1.3e-4} \\  
\bottomrule
\end{tabular}
\label{tab:comparisonSWE}
\end{minipage}
\hspace{1em}
\begin{minipage}{0.45\textwidth}
\centering
        \caption{Results on  CloudCast dataset.}
        \begin{tabular}{lcc}
            \toprule
            Method & SSIM ($\uparrow)$ & PSNR ($\uparrow$) \\
            \midrule
            AE-ConvLSTM \cite{zhang2017deep} & \second{0.66} & \second{8.06} \\
            MD-GAN \cite{xiong2018learning} & \third{0.60} & \third{7.83} \\
            TVL1 \cite{urbich2018novel} & 0.58 & 7.50 \\
            Persistent \cite{zhang2017deep} & 0.55 & 7.41 \\
            \midrule
            {ADRNet} & \first{0.83} & \first{38.17} \\
            \bottomrule
        \end{tabular}
        \label{tab:cloudcast}
\end{minipage}
\end{table}

\begin{figure}[h] 
\centering
\begin{tabular}{ccc}
\includegraphics[width=4.0cm]{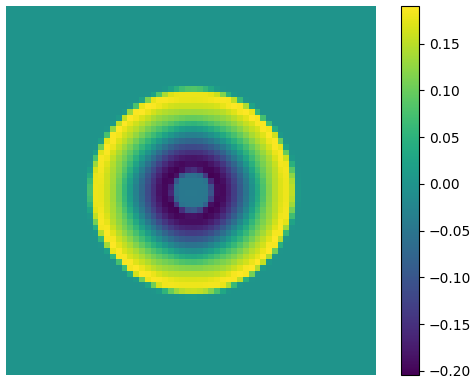} &
\includegraphics[width=4.0cm]{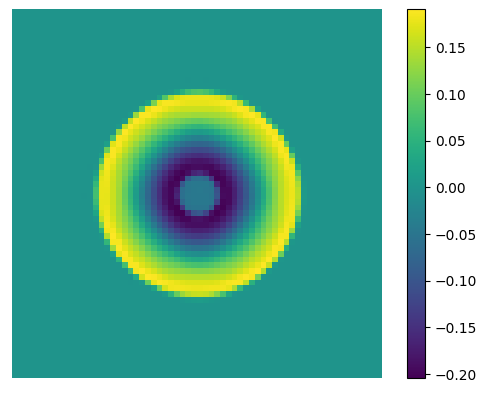} &
\includegraphics[width=4.1cm]{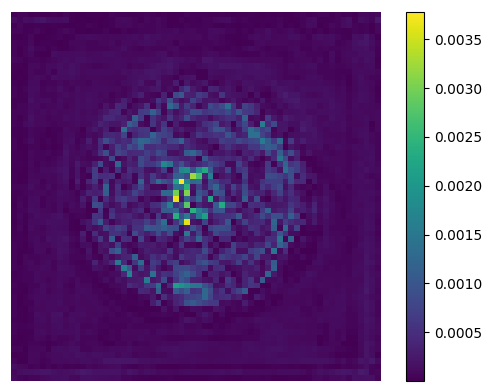} \\
Ground-Truth & Prediction & Error
\end{tabular}
\caption{Prediction and error for the SWE problem \label{fig:swe} using our ADRNet.}
\end{figure}

\begin{figure}[ht]
    \centering
    
    \begin{minipage}{0.05\textwidth}
        \rotatebox{90}{\scalebox{0.7}{\textbf{Ground Truth}}}
        
    \end{minipage}%
    \begin{minipage}{0.18\textwidth}
        \hspace{-5mm}
        \includegraphics[width=\linewidth]{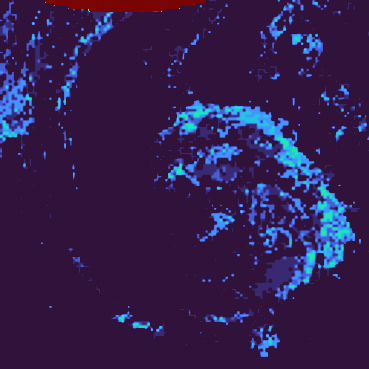}
        \centering
        \scalebox{0.7}{$t+1$}
    \end{minipage}%
    \begin{minipage}{0.18\textwidth}
    \hspace{-5mm}
        \includegraphics[width=\linewidth]{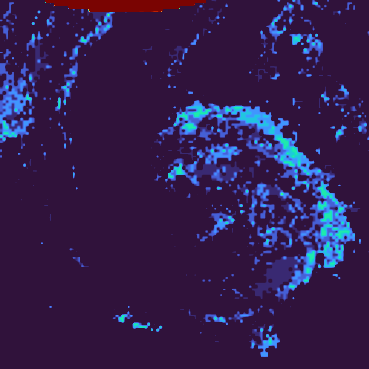}
        \centering
        \scalebox{0.7}{$t+2$}
    \end{minipage}%
    \begin{minipage}{0.18\textwidth}
    \hspace{-5mm}
        \includegraphics[width=\linewidth]{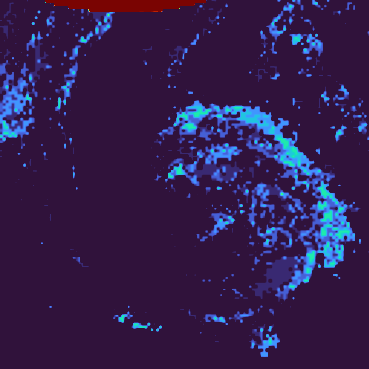}
        \centering
        \scalebox{0.7}{$t+3$}
    \end{minipage}%
    \begin{minipage}{0.18\textwidth}
    \hspace{-5mm}
        \includegraphics[width=\linewidth]{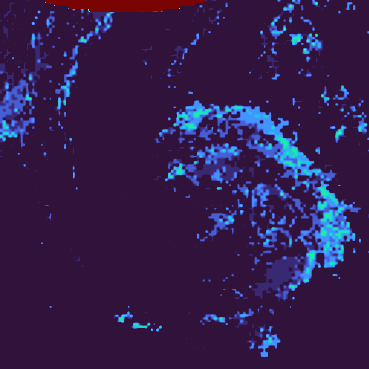}
        \centering
        \scalebox{0.7}{$t+4$}
    \end{minipage}%
    \hspace{1mm}
    \begin{minipage}{0.03\textwidth}
    \vspace{-3mm}
         \includegraphics[width=0.65cm]{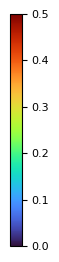}
    \end{minipage}

    \vspace{1mm} 

    \begin{minipage}{0.05\textwidth}
        \rotatebox{90}{\scalebox{0.7}{\textbf{Prediction}}}
    \end{minipage}%
    \begin{minipage}{0.18\textwidth}
    \hspace{-5mm}
        \includegraphics[width=\linewidth]{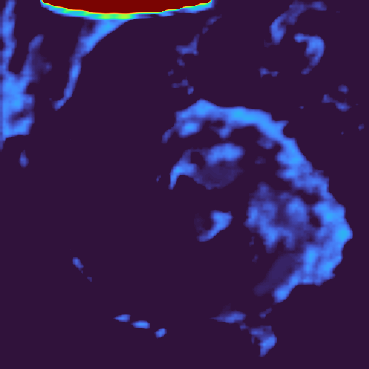}
        \centering
        \scalebox{0.6}{$t+1$}
    \end{minipage}%
    \begin{minipage}{0.18\textwidth}
    \hspace{-5mm}
        \includegraphics[width=\linewidth]{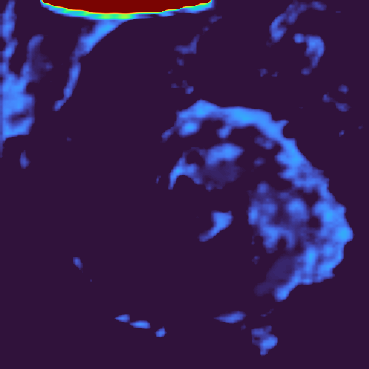}
        \centering
        \scalebox{0.6}{$t+2$}
    \end{minipage}%
    \begin{minipage}{0.18\textwidth}
    \hspace{-5mm}
        \includegraphics[width=\linewidth]{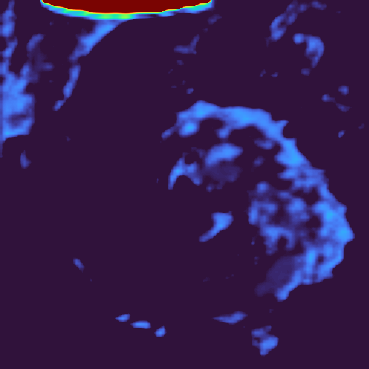}
        \centering
        \scalebox{0.6}{$t+3$}
    \end{minipage}%
    \begin{minipage}{0.18\textwidth}
    \hspace{-5mm}
        \includegraphics[width=\linewidth]{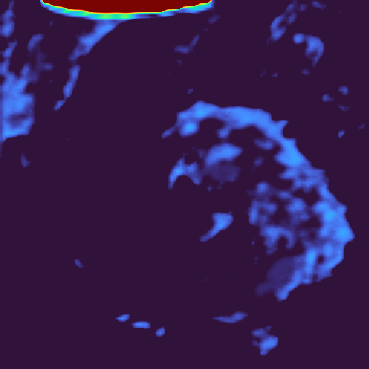}
        \centering
        \scalebox{0.6}{$t+4$}
    \end{minipage}%
    \hspace{1mm}
    \begin{minipage}{0.03\textwidth}
    \vspace{-3mm}
        \includegraphics[width=0.65cm]{Img/Output_Images_Colorbar.png}
    \end{minipage}

    \vspace{1mm} 

    \begin{minipage}{0.05\textwidth}
        \rotatebox{90}{\scalebox{0.7}{\textbf{Difference}}}
    \end{minipage}%
    \begin{minipage}{0.18\textwidth}
    \hspace{-5mm}
        \includegraphics[width=\linewidth]{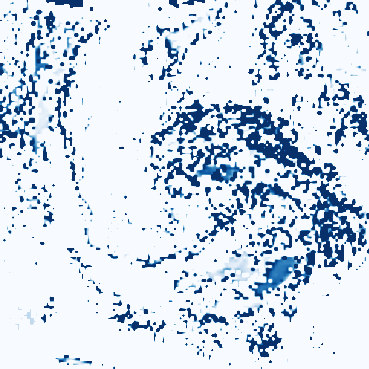}
        \centering
        \scalebox{0.7}{$t+1$}
    \end{minipage}%
    \begin{minipage}{0.18\textwidth}
    \hspace{-5mm}
        \includegraphics[width=\linewidth]{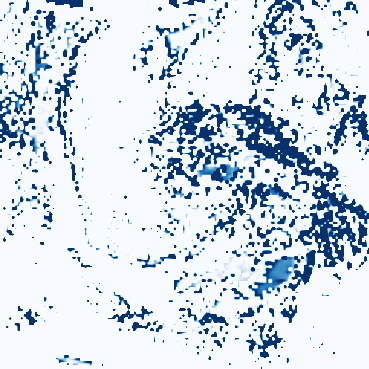}
        \centering
        \scalebox{0.7}{$t+2$}
    \end{minipage}%
    \begin{minipage}{0.18\textwidth}
    \hspace{-5mm}
        \includegraphics[width=\linewidth]{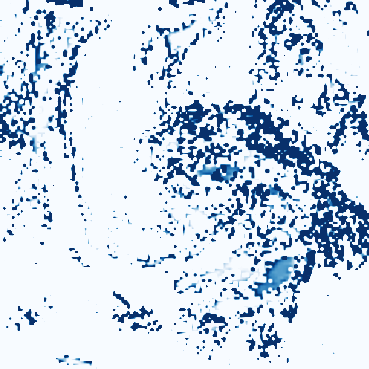}
        \centering
        \scalebox{0.7}{$t+3$}
    \end{minipage}%
    \begin{minipage}{0.18\textwidth}
    \hspace{-5mm}
        \includegraphics[width=\linewidth]{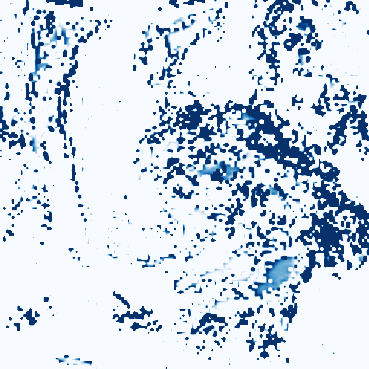}
        \centering
        \scalebox{0.7}{$t+4$}
    \end{minipage}%
    \hspace{1mm}
    \begin{minipage}{0.03\textwidth}
    \vspace{-3mm}
        \includegraphics[width=1cm]{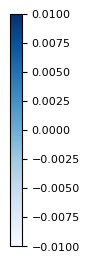}
    \end{minipage}

    \caption{Example of the forecast for ADRNet relative to ground truth for four time steps.}
    \label{fig:cloudcast_performance}
\end{figure}

\ref{fig:swe} and  Figure \ref{fig:cloudcast_performance}.
Examples of the predictions of the SWE dataset and the CloudCast datasets are plotted in Figure
For the SWE dataset, the errors are very small and close to machine precision. For CloudCast, the data is noisy, and it is not clear how well it should fit.
Predicting a single-time step, while useful, has limited applicability. Our goal is to push the prediction for longer, hence providing an alternative to expensive integration. The results for SWE for long-time prediction are presented in Table~\ref{tab:swe_long}, together with a comparison of the FNO method \cite{li2020fourier} where we see that our model performs well even for long-time prediction.
\begin{table}[h]
\footnotesize
\centering
\caption{Comparison of ADRNet and FNO across Long-Range Predictions.}
\label{tab:swe_long}
\begin{tabular}{@{}lccccccccc@{}}
\toprule
\textbf{Metric} & \multicolumn{2}{c}{\textbf{10 $\rightarrow$ 5}} & \multicolumn{2}{c}{\textbf{10 $\rightarrow$ 10}} & \multicolumn{2}{c}{\textbf{10 $\rightarrow$ 20}} & \multicolumn{2}{c}{\textbf{10 $\rightarrow$ 50}} \\ 
\cmidrule(lr){2-3} \cmidrule(lr){4-5} \cmidrule(lr){6-7} \cmidrule(lr){8-9}
 & \textbf{ADRNet} & \textbf{FNO} & \textbf{ADRNet} & \textbf{FNO} & \textbf{ADRNet} & \textbf{FNO} & \textbf{ADRNet} & \textbf{FNO} \\ \midrule
MSE $\downarrow$ & 9.2e-08 & 4.0e-07 & 1.5e-07 & 5.8e-07 & 2.1e-07 & 6.7e-07 & 8.5e-07 & 1.4e-06 \\ 
nMSE $\downarrow$ & 8.5e-08 & 3.7e-07 & 1.4e-07 & 5.4e-07 & 1.9e-07 & 6.2e-07 & 7.8e-07 & 1.3e-06 \\ 
RMSE $\downarrow$ & 3.0e-04 & 6.3e-04 & 3.9e-04 & 7.6e-04 & 4.5e-04 & 8.1e-04 & 9.2e-04 & 1.2e-03 \\ 
nRMSE $\downarrow$ & 2.9e-04 & 6.1e-04 & 3.7e-04 & 7.3e-04 & 4.4e-04 & 7.8e-04 & 8.8e-04 & 1.1e-03 \\
MAE $\downarrow$ & 2.0e-04 & 2.8e-04 & 1.7e-04 & 3.7e-04 & 1.9e-04 & 3.6e-04 & 4.1e-04 & 5.7e-04 \\ \bottomrule
\end{tabular}
\end{table}

\paragraph{Video Prediction Performance.}
We have used a number of video datasets to test our ADRNet. The results of two of them (Moving MNIST and KITTI) are reported in Table \ref{tab:comparisonMNIST} and Table \ref{tab:comparisonKITTI}. We perform additional experiments for the KTH Action and TaxiBJ datasets in the appendix \ref{app:exp}. The moving MNIST dataset adheres to the assumptions of our ADRNet. Indeed, for this dataset, we obtain results that are very close to state-of-the-art methods.
\begin{table}[h]
\setlength{\tabcolsep}{0.3em}
\footnotesize
\centering

\begin{minipage}{0.3\textwidth}    
    \centering
    \caption{Moving MNIST.}
    \begin{tabular}[b]{lcc}
    
\toprule
Method & MSE $\downarrow$ & MAE $\downarrow$   \\ 
\midrule
MSPred \cite{VillarCorrales2022MSPredVP} & 34.4 & -  \\
MAU \cite{NEURIPS2021_e25cfa90} & 27.6 & -  \\
PhyDNet \cite{Guen2020DisentanglingPD} & 24.4 & 70.3  \\
SimVP \cite{tan2022simvp} & 23.8 & 68.9  \\
CrevNet \cite{Yu2020Efficient} & 22.3 & -  \\
TAU \cite{tan2023temporal} & 19.8 & \third{60.3}  \\
SwinLSTM \cite{tang2023swinlstm} & \third{17.7} & -  \\
IAM4VP \cite{seo2023implicit} & \first{15.3} & \first{49.2}  \\
\midrule
{ADRNet} & \second{16.1} & \second{50.3}  \\
\bottomrule
\end{tabular}
\label{tab:comparisonMNIST}
\end{minipage}%
\hspace{3.5em}
\begin{minipage}{0.5\textwidth} 
    \centering
        \caption{Results on KITTI.}
    \begin{tabular}[t]{lcccc}
        \toprule
        \textbf{Method} & \multicolumn{2}{c}{\textbf{MS-SSIM ($\times 10^{-2}$) $\uparrow$}} & \multicolumn{2}{c}{\textbf{LPIPS ($\times 10^{-2}$) $\downarrow$}} \\
        \cmidrule(r){2-3} \cmidrule(r){4-5}
        & $t+1$ & $t+3$ & $t+1$ & $t+3$ \\
        \midrule
        SADM \cite{bei2021learning} & 83.06 & 72.44 & 14.41 & 24.58 \\
        MCNET \cite{villegas2017decomposing} & 75.35 & 63.52 & 24.04 & 37.71 \\
        CorrWise \cite{geng2022comparing} & 82.00 & N/A & 17.20 & N/A \\
        OPT \cite{wu2022optimizing} & 82.71 & 69.50 & 12.34 & 20.29 \\
        DMVFN \cite{hu2023dynamic}  & \second{88.06} & \third{76.53} & \third{10.70} & \third{19.28} \\
        DMVFN \cite{hu2023dynamic} & \first{88.53} & \second{78.01} & \third{10.74} & \second{19.27} \\
        \midrule
        {ADRNet} &  \third{85.86} & \first{83.62} & \first{7.54} & \first{9.26} \\
        \bottomrule
        \label{tab:comparisonKITTI}
    \end{tabular}
\end{minipage}
\end{table}

The KTH Action dataset is more complex as not all frames can be predicted from the previous frames without generation power. Nonetheless, even for this dataset our ADRNet performs close to the state of the art. This limiting aspect of video synthesis is studied through experiments in appendix \ref{limiting_datasets}.

\section{Conclusion}
In this paper, we have presented a new network for tasks that reside on a regular mesh that can be viewed as a multi-channel image. The method combines standard convolutions with a linear operator that transports information from one part of the image to another. The transportation vector field is learned from
previous images (that is, history), allowing for 
information to pass from different parts of the image to others without loss.
We combine this information within a diffusion-reaction process that can be coded by itself or by using a standard ResNet.

\newpage
{
\bibliographystyle{plain}
\bibliography{biblio}
}


\appendix

\clearpage
\section{Ablation Studies and Additional Experiments}
\label{app:exp}

\subsection{Limited Generative Synthesis}
\label{limiting_datasets}
\begin{figure}[h] 
\centering
\begin{tabular}{cc}
\includegraphics[width=7cm]{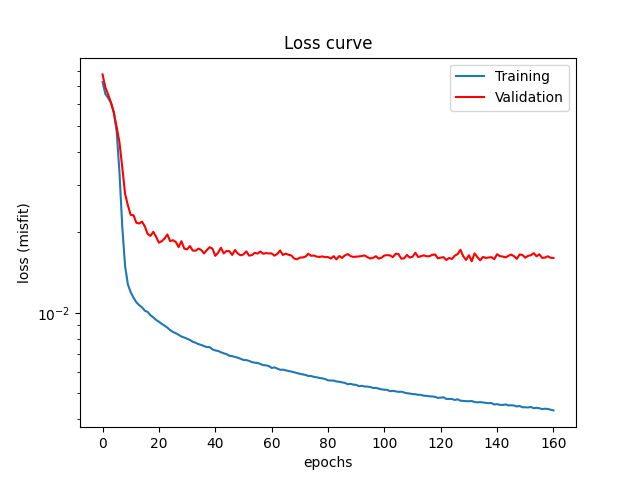} &
\includegraphics[width=7cm]{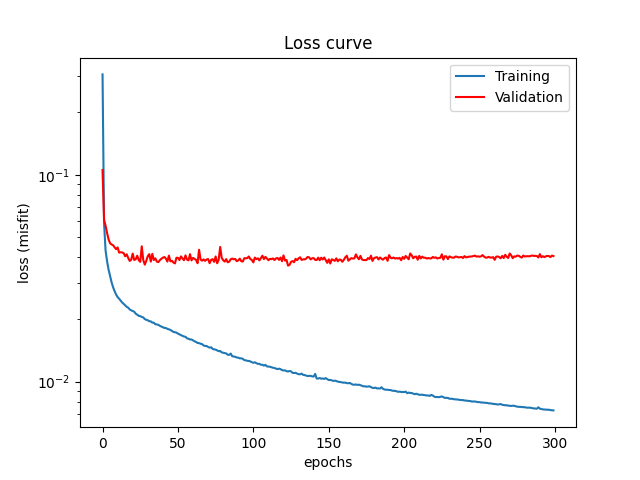} \\
KTH Action & TaxiBJ\\
\end{tabular}
\caption{Bias in training and testing samples in KTH Action and TaxiBJ datasets }
\label{fig:kth_taxi}
\end{figure}

Two real-life video datasets are taken to predict future time frames. Their statistics can be found in Table \ref{tab:additionaldatasets}. The specific challenge posed by these datasets is due to the dissimilarity in the train and test sets. This is evident from the notable difference in the training and validation/test losses, which can be seen in Figure \ref{fig:kth_taxi}. The validation loss starts increasing with more epochs. For example, the KTH Action uses the movement behavior of 16 people for training while the models are tested on the movement behavior of 9 other people in a slightly altered scenario. So, we can say that the problem is to learn the general logic to predict unseen scenarios. Thus generative capability of a model could be crucial for better prediction.  

\paragraph{KTH Action}
The KTH dataset features 25 individuals executing six types of actions: walking, jogging, running, boxing, hand waving, and hand clapping. Following methodologies established in references \cite{villegas2017decomposing,wang2018eidetic}, we utilize individuals 1-16 for training and individuals 17-25 for testing. The models are trained to predict the subsequent 20 frames based on the preceding 10 observations.

\paragraph{TaxiBJ}
TaxiBJ is a collection of real-world GPS spatiotemporal data of taxis recorded as frames of 32x32x2 heat maps every half an hour, quantifying traffic flow in Beijing. We split the whole dataset into a training set and a test set as described in \cite{zhang2017deep}. We train the networks to predict 4 future time frames from 4 observations. 

\paragraph{Results}
Our model is easily able to predict and outperform the state of art models in real-life video examples as well. The results for KTH Action and TaxiBJ can be seen in \ref{tab:comparisonKTH} and \ref{tab:comparisonTaxiBJ}. 

\begin{table}[ht]
\centering
\caption{Additional Dataset Statistics: Details on Training and Testing, Image Sequences, and Resolutions}
\vspace{10pt}
\begin{tabular}{lccccc}
\toprule
\textbf{Dataset} & \( N_{\text{train}} \) & \( N_{\text{test}} \) & \( (C, H, W) \) & \( History \) & \( Prediction \) \\
\midrule
KTH Action        & 5200                  & 3167                 & (1, 128, 128)     & 10       & 20        \\
TaxiBJ        & 19627                  & 1334                 & (2, 32, 32)     & 4       & 4        \\
\bottomrule
\end{tabular}

\label{tab:additionaldatasets}
\end{table}

\begin{table}[h]
\centering
\caption{Comparison of Our Method for KTH Action Dataset.}
\label{tab:comparisonKTH}
\begin{tabular}{lcccccc}
\toprule
Method & SSIM $\uparrow$ & PSNR (dB) $\uparrow$\\
\midrule

ConvLSTM \cite{convLSTM} & 0.712 & 23.58\\
PredRNN \cite{NIPS2017_e5f6ad6c} & 0.839 & 27.55\\
CausalLSTM \cite{pmlr-v80-wang18b} & 0.865 & 28.47\\
MSPred \cite{VillarCorrales2022MSPredVP} & {0.930} & 28.93\\
E3D-LSTM \cite{wang2018eidetic} & 0.879 & 29.31\\
SimVP \cite{tan2022simvp} & {0.905} & {33.72}\\    
TAU \cite{tan2023temporal} & {0.911} & {34.13}\\  
SwinLSTM \cite{tang2023swinlstm} & {0.903} &  {34.34}\\
\midrule
{ADRNet} & {0.808} & {31.58}\\
\bottomrule
\end{tabular}
\end{table}

\begin{table}[h]
\centering
\caption{Comparison of Our Method for TaxiBJ Dataset.}
\label{tab:comparisonTaxiBJ}
\begin{tabular}{lcccccc}
\toprule
Method & MSE $\downarrow$ & MAE $\downarrow$ & SSIM $\uparrow$ \\
\midrule
ST-ResNet \cite{zhang2017deep} & 0.616 & - & - \\
VPN \cite{pmlr-v70-kalchbrenner17a} & 0.585 & - & - \\
ConvLSTM\cite{convLSTM} & 0.485 & 17.7 & 0.978 \\
FRNN \cite{oliu2018folded} & 0.482 & - & - \\
PredRNN \cite{NIPS2017_e5f6ad6c} & 0.464 & 17.1 & 0.971 \\
CausalLSTM \cite{pmlr-v80-wang18b} & 0.448 & 16.9 & 0.977 \\
MIM \cite{wang2019memory} & 0.429 & 16.6 & 0.971 \\
E3D-LSTM \cite{wang2018eidetic} & 0.432 & 16.9 & 0.979 \\
PhyDNet \cite{Guen2020DisentanglingPD} & 0.419 & 16.2 & 0.982 \\
SimVP \cite{tan2022simvp} & 0.414 & 16.2 & 0.982 \\
SwinLSTM \cite{tang2023swinlstm} & 0.390 & - & 0.980 \\
IAM4VP \cite{seo2023implicit} & 0.372 & 16.4 & 0.983 \\
TAU \cite{tan2023temporal} & 0.344 & 15.6 & 0.983 \\
\midrule
{ADRNet} & {0.445} & {16.6} & {0.975} \\
\bottomrule
\end{tabular}
\end{table}

\subsection{CloudCast}
The CloudCast dataset is used for multiple long-range predictions like 4, 8, 12, and 16 timesteps. It can be noticed that even if the MSE or the quality degrades, the degradation is noticeably minimal. It can be seen in Table \ref{ccast}, the figures for predicting 16 steps in future is still better than the state of art for 4 steps in future.  

\begin{table}[H]
    \centering
    \caption{Results for CloudCast dataset. Comparison of our model (ADRNet) with  state of art models}
    \begin{minipage}{0.5\textwidth}
        \centering
    \end{minipage}%
    \begin{minipage}{0.5\textwidth}
        \centering
        \label{ccast}
        \begin{tabular}{lcccc}
            \toprule
            \multicolumn{5}{c}{ADRNet Predictive performance} \\
            \midrule
            Metric & t + 4 & t + 8 & t + 12 & t + 16 \\
            \midrule
            MSE ($\downarrow$) & 0.015 & 0.016 &  0.018 &  0.019\\
            SSIM ($\uparrow$) & 0.83 & 0.79 &  0.76 &  0.74\\
            PSNR ($\uparrow$) & 38.17 & 37.89 & 37.35  & 37.23\\
            \bottomrule
        \end{tabular}
    \end{minipage}
\end{table}

\section{Evaluation Metrics}
\label{sec8}
\paragraph{Moving MNIST, KTH Action, TaxiBJ, CloudCast}
These specific video prediction datasets have been using MAE (Mean Absolute Error), MSE (Mean Squared Error), SSIM (Structural Similarity) and PSNR (Peak Signal-to-Noise Ratio). The evaluated SSIM and PSNR are averaged over each image. The MSE and MAE have a specific way to calculate, where the pixel-wise evaluation values are summed up for all the pixels in the image.  

\begin{equation}
\text{MSE} = \frac{1}{N} \sum_{i=1}^{N} \sum_{h=1}^{H} \sum_{w=1}^{W} \sum_{c=1}^{C} (y - \hat{y})^2
\end{equation}
\begin{equation}
\text{MAE} = \frac{1}{N} \sum_{i=1}^{N} \sum_{h=1}^{H} \sum_{w=1}^{W} \sum_{c=1}^{C} |y - \hat{y}|
\end{equation}
\begin{equation}
\text{PSNR} =  \frac{1}{N} \sum_{i=1}^{N} 10 \cdot \log_{10} \left( \frac{\text{MAX}^2}{\text{MSE}} \right)
\end{equation}
\begin{equation}
\text{SSIM(x,y)} = \frac{(2 \mu_x \mu_y + C_1)(2 \sigma_{xy} + C_2)}{(\mu_x^2 + \mu_y^2 + C_1)(\sigma_x^2 + \sigma_y^2 + C_2)}  
\end{equation}
\begin{equation}
\overline{\text{SSIM}} = \frac{1}{N} \sum_{i=1}^{N} \text{SSIM}(x,y)
\end{equation}

where:
\begin{align*}
N & \text{ is the number of images in the dataset,} \\
H & \text{ is the height of the images,} \\
W & \text{ is the width of the images,} \\
C & \text{ is the number of channels (e.g., 3 for RGB images),} \\
y & \text{ is the true pixel value at position } (i, h, w, c), \text{ and} \\
\hat{y} & \text{ is the predicted pixel value at position } (i, h, w, c).\\
\text{MAX} & \text{ is the maximum possible pixel value of the image (e.g., 255 for an 8-bit image),} \\
\text{MSE} & \text{ is the Mean Squared Error between the original and compressed image.}\\
\mu_x & \text{ is the average of } x, \\
\mu_y & \text{ is the average of } y, \\
\sigma_x^2 & \text{ is the variance of } x, \\
\sigma_y^2 & \text{ is the variance of } y, \\
\sigma_{xy} & \text{ is the covariance of } x \text{ and } y, \\
C_1 & = (K_1 L)^2 \text{ and } C_2 = (K_2 L)^2 \text{ are two variables to stabilize the division with weak denominator,} \\
L & \text{ is the dynamic range of the pixel values (typically, this is 255 for 8-bit images),} \\
K_1 & \text{ and } K_2 \text{ are small constants (typically, } K_1 = 0.01 \text{ and } K_2 = 0.03).\\
\end{align*}

\paragraph{PDEBench-SWE}
PDEBench uses the concept of pixel-wise mean squared error (MSE) and normalized mean squared error (nMSE) to validate scaled variables in simulated PDEs. Along with these, we also use root mean squared error (RMSE) and normalized root mean squared error (nRMSE).

\begin{equation}
\text{MSE} = \frac{1}{N \cdot H \cdot W \cdot C} \sum_{n=1}^{N} \sum_{h=1}^{H} \sum_{w=1}^{W} \sum_{c=1}^{C} (x - \hat{x})^2
\end{equation}
\begin{equation}
\text{nMSE} = \frac{1}{N \cdot H \cdot W \cdot C} \sum_{n=1}^{N} \sum_{h=1}^{H} \sum_{w=1}^{W} \sum_{c=1}^{C} \frac{(x - \hat{x})^2}{x^2}
\end{equation}
\begin{equation}
\text{RMSE} = \frac{1}{N}\sqrt{\frac{1}{ H \cdot W \cdot C} \sum_{n=1}^{N} \sum_{h=1}^{H} \sum_{w=1}^{W} \sum_{c=1}^{C} (x - \hat{x})^2}
\end{equation}
\begin{equation}
\text{nRMSE} = \frac{1}{N}\sqrt{\frac{1}{ H \cdot W \cdot C} \sum_{n=1}^{N} \sum_{h=1}^{H} \sum_{w=1}^{W} \sum_{c=1}^{C} \frac{(x - \hat{x})^2}{x^2}}
\end{equation}

where:
\begin{align*}
N & \text{ is the number of images in the dataset,} \\
H & \text{ is the height of the images,} \\
W & \text{ is the width of the images,} \\
C & \text{ is the number of channels (e.g., 3 for RGB images),} \\
x & \text{ is the true pixel value at position } (n, h, w, c), \text{ and} \\
\hat{x} & \text{ is the predicted pixel value at position } (n, h, w, c).
\end{align*}

\paragraph{KITTI}

\begin{equation}
\text{MS-SSIM}(x, y) = \left[ l_M(x, y) \right]^{\alpha_M} \prod_{j=1}^{M} \left[ c_j(x, y) \right]^{\beta_j} \left[ s_j(x, y) \right]^{\gamma_j}
\end{equation}
\begin{equation}
\overline{\text{MS-SSIM}} = \frac{1}{N} \sum_{i=1}^{N} \text{MS-SSIM}(x,y)
\end{equation}

where:
\begin{align*}
N & \text{ is the number of images in the dataset,} \\
l_M(x, y) & \text{ is the luminance comparison at the coarsest scale } M, \\
c_j(x, y) & \text{ is the contrast comparison at scale } j, \\
s_j(x, y) & \text{ is the structure comparison at scale } j, \\
\alpha_M, \beta_j, \gamma_j & \text{ are the weights applied to the luminance, contrast, and structure terms at each scale respectively,} \\
M & \text{ is the number of scales used in the comparison.}
\end{align*}

The luminance, contrast, and structure comparisons are given by:

\begin{align*}
l(x, y) &= \frac{2 \mu_x \mu_y + C_1}{\mu_x^2 + \mu_y^2 + C_1} \\
c(x, y) &= \frac{2 \sigma_x \sigma_y + C_2}{\sigma_x^2 + \sigma_y^2 + C_2} \\
s(x, y) &= \frac{\sigma_{xy} + C_3}{\sigma_x \sigma_y + C_3}
\end{align*}

where:
\begin{align*}
\mu_x, \mu_y & \text{ are the local means of } x \text{ and } y, \\
\sigma_x, \sigma_y & \text{ are the local standard deviations of } x \text{ and } y, \\
\sigma_{xy} & \text{ is the local covariance of } x \text{ and } y, \\
C_1, C_2, C_3 & \text{ are constants to stabilize the division.}
\end{align*}

\begin{equation}
\text{LPIPS}(x, \hat{x}) = \sum_{l} \frac{1}{H_l W_l} \sum_{h=1}^{H_l} \sum_{w=1}^{W_l} \| w_l \odot (\phi_l(x)_{hw} - \phi_l(\hat{x})_{hw}) \|_2^2
\end{equation}
\begin{equation}
\overline{\text{LPIPS}} = \frac{1}{N} \sum_{i=1}^{N} \text{LPIPS}(x,\hat{x})
\end{equation}

where:
\begin{align*}
N & \text{ is the number of images in the dataset,} \\
\phi_l(x) & \text{ is the activation of the } l\text{-th layer of a deep network for the image } x, \\
\phi_l(\hat{x}) & \text{ is the activation of the } l\text{-th layer of a deep network for the image } \hat{x}, \\
w_l & \text{ is a learned weight vector for the } l\text{-th layer,} \\
H_l & \text{ and } W_l \text{ are the height and width of the } l\text{-th layer activations,} \\
\odot & \text{ denotes element-wise multiplication.}
\end{align*}

\section{Hyperparameter Settings and Computational Resources}
\subsection{ADRNet Training on PDEBench-SWE}
\label{settings}
\begin{table}[h]
\centering
\begin{tabular}{lcccc}
\toprule
\textbf{Hyperparameter}     & \textbf{Symbol} & \textbf{Value}  \\ \midrule
Learning Rate               & $\eta$          & $1e-04$         \\
Batch Size                  & $B$             & $64$           \\
Number of Epochs            & $N$             & $200$           \\
Optimizer                   & -               & Adam            \\
Number of Layers            & -               & $1$             \\
Hidden Channels      & -               & $128$            \\
Activation Function         & -               & SiLU            \\ 
\bottomrule
\end{tabular}
\caption{Neural Network Hyperparameters}
\label{tab:hyperparameters_swe}
\end{table}

\subsection{ADRNet Training on Other Datasets}
\label{settings2}
\begin{table}[h]
\centering
\begin{tabular}{lcc}
\toprule
\textbf{Hyperparameter}     & \textbf{Symbol} & \textbf{Value}  \\ \midrule
Learning Rate               & $\eta$          & $2e-06$         \\
Batch Size                  & $B$             & $16$           \\
Number of Epochs            & $N$             & $1000$           \\
Optimizer                   & -               & Adam            \\
Number of Layers            & -               & $8$             \\
Hidden Channels      & -               & $192$            \\
Activation Function         & -               & SiLU            \\
Learning Rate Scheduler     & -               & ExponentialLR      \\ \bottomrule
\end{tabular}
\caption{Neural Network Hyperparameters}
\label{tab:hyperparameters}
\end{table}

\subsection{Computational Resources} 
\label{resources}
All our experiments are conducted using an NVIDIA RTX-A6000 GPU with 48GB of memory.

\end{document}